\documentclass[11pt,a4paper]{article}
\usepackage[hyperref]{acl2018}
\usepackage{times}
\usepackage{latexsym}
\usepackage{url}

\aclfinalcopy
\def\aclpaperid{1335}

\usepackage{booktabs}
\usepackage{graphicx}
\usepackage[export]{adjustbox}
\usepackage{xspace}
\usepackage{amsmath,amssymb}
\usepackage{dsfont}
\usepackage[skip=5pt]{caption}
\graphicspath{ {images/} }

\usepackage[hyperref]{acl2018}
\hypersetup{draft}

\usepackage{color}

\newcommand{\lmtt}[1]{\fontfamily{lmtt}\selectfont{#1}}

\newcommand{\bl}{{\lmtt BabbleLabble}\xspace}
\newcommand{\spouse}{{\lmtt Spouse}\xspace}
\newcommand{\disease}{{\lmtt Disease}\xspace}
\newcommand{\protein}{{\lmtt Protein}\xspace}

\newcommand{\sX}{\mathcal X}

\DeclareMathOperator*{\argmin}{arg\,min}

\title{Training Classifiers with Natural Language Explanations}

\author{
  Braden Hancock \\
  Computer Science Dept. \\
  Stanford University \\
  {\tt \small{bradenjh@cs.stanford.edu}} \\
  \And
  Paroma Varma \\
  Electrical Engineering Dept. \\
  Stanford University \\
  {\tt \small{paroma@stanford.edu}} \\
  \And
  Stephanie Wang \\
  Computer Science Dept. \\
  Stanford University \\
  {\tt \small{steph17@stanford.edu}} \\
  \AND
  Martin Bringmann \\
  OccamzRazor \\
  San Francisco, CA \\
  {\tt \small{martin@occamzrazor.com}} \\
  \And
  Percy Liang \\
  Computer Science Dept. \\
  Stanford University \\
  {\tt \small{pliang@cs.stanford.edu}} \\
  \And
  Christopher R{\'e} \\
  Computer Science Dept. \\
  Stanford University \\
  {\tt \small{chrismre@cs.stanford.edu}} \\
}

\date{8 May 2018}

\begin{document}
\maketitle

\begin{abstract}
Training accurate classifiers requires many labels,
but each label provides only \mbox{limited} information (one bit for binary \mbox{classification}).
In this work, we \mbox{propose} \bl, a framework for training classifiers
in which an \mbox{annotator} \mbox{provides} a natural language explanation for each labeling decision.
A semantic parser converts these explanations into programmatic labeling functions
that generate noisy labels for an arbitrary amount of \mbox{unlabeled} data,
which is used to train a classifier.
On three relation \mbox{extraction} tasks, we find that users are able to train classifiers with comparable F1 scores from 5--100$\times$ faster by \mbox{providing} explanations instead of just \mbox{labels}.
\mbox{Furthermore}, given the inherent imperfection of labeling \mbox{functions},
  we find that a simple rule-based semantic parser suffices.
\end{abstract}

\begin{figure}[t]
  \centering
  \includegraphics[width=3in]{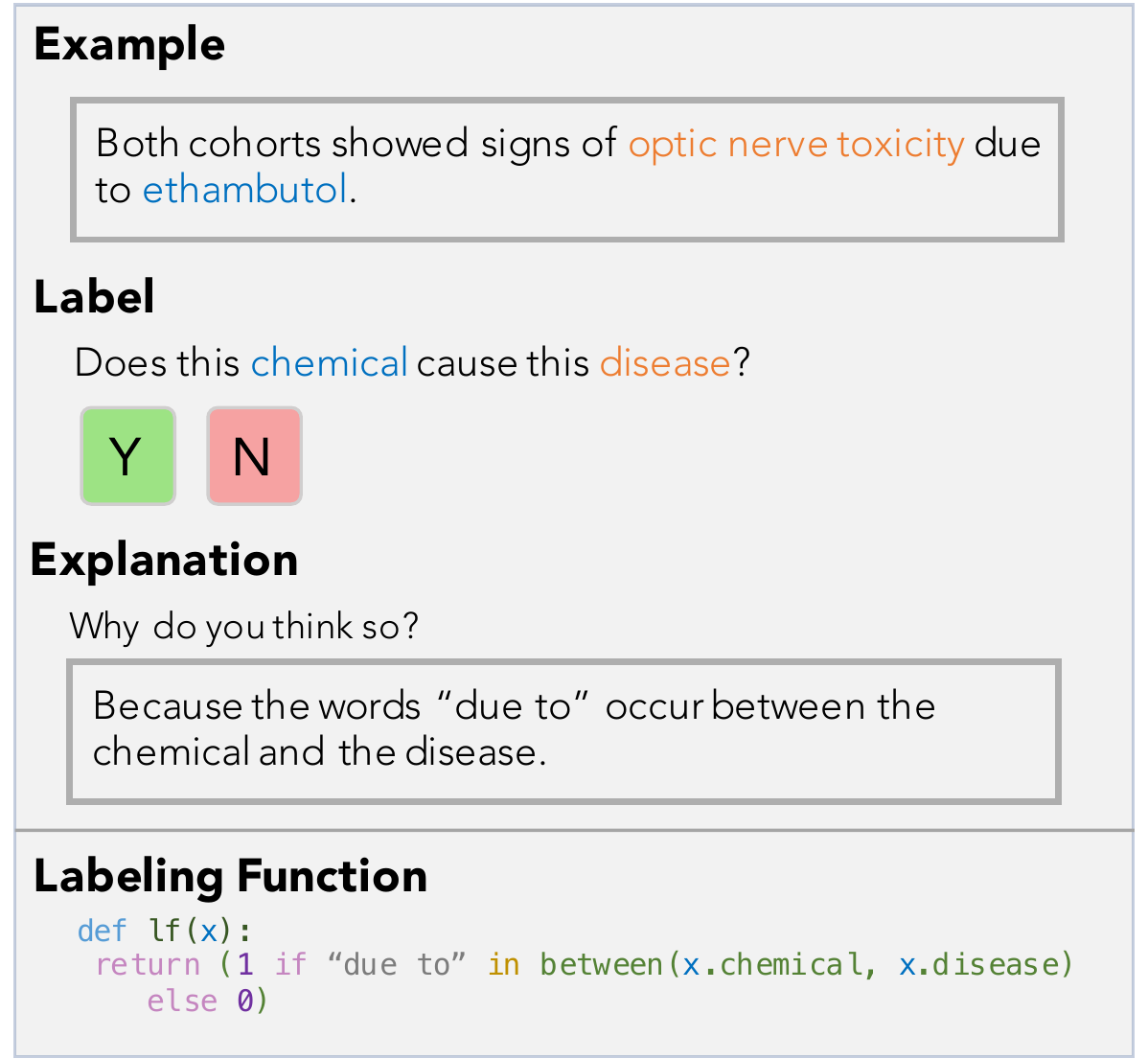}
  \caption{In \bl, the user provides a natural language explanation for each labeling \mbox{decision}. These explanations are parsed into \mbox{labeling} functions that convert unlabeled data into a large labeled dataset for training a classifier.}
  \label{fig:interface}
\end{figure}

\section{Introduction}

The standard protocol for obtaining a labeled dataset is to have a human annotator view each example, assess its relevance, and provide a label (e.g., positive or negative for binary classification).
However, this only provides one bit of information per example.
This invites the question: how can we get more information per example, given that the annotator has already spent the effort reading and understanding an example?

\begin{figure*}[t]
  \centering
  \includegraphics[width=6in]{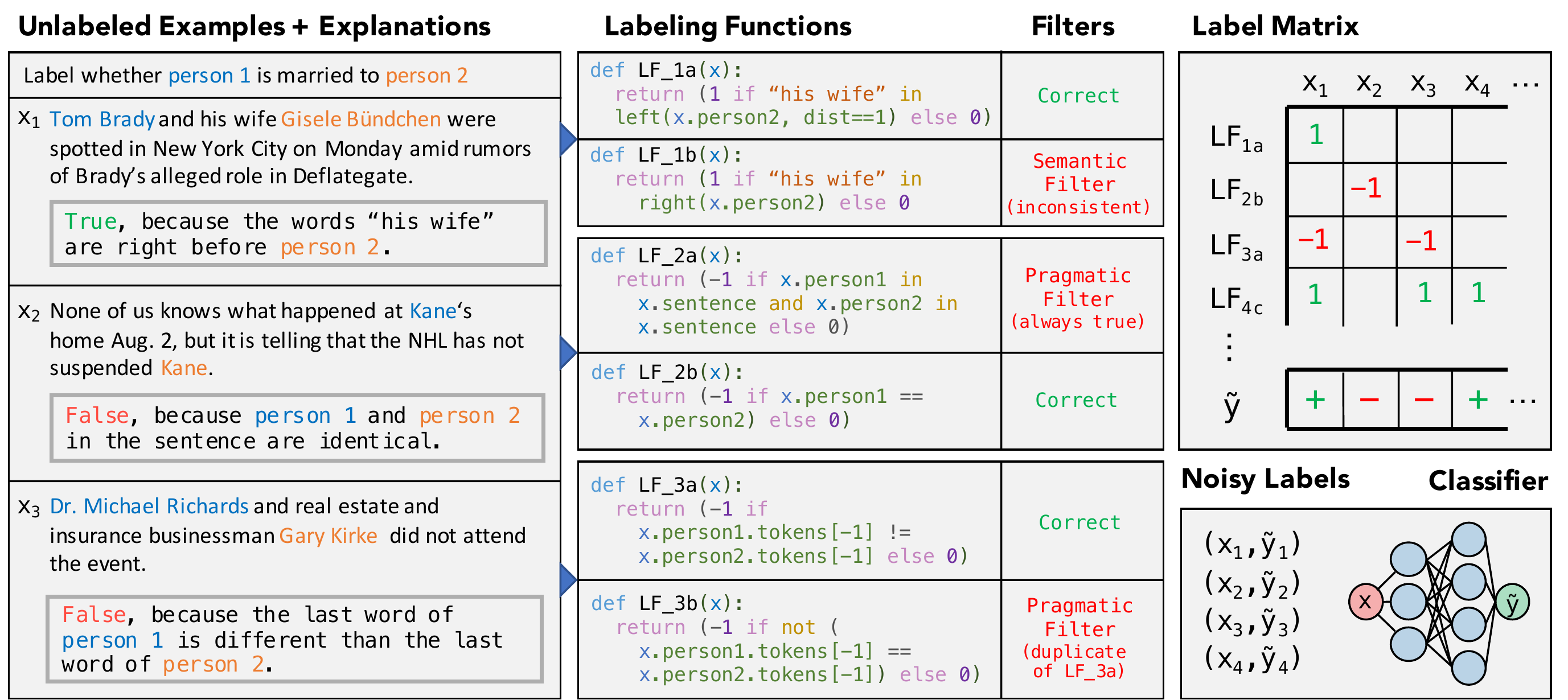}
  \caption{
    Natural language explanations are parsed into candidate labeling functions (LFs).
    Many \mbox{incorrect} LFs are filtered out automatically by the filter bank.
    The remaining functions provide heuristic labels over the unlabeled dataset, which are aggregated into one noisy label per example, yielding a large, noisily-labeled training set for a classifier.}
  \label{fig:walkthrough}
\end{figure*}

Previous works have relied on identifying relevant parts of the input such as
labeling features \citep{druck2009active, raghavan2005interactive, liang09measurements},
highlighting rationale phrases in text \citep{zaidan08annotator, arora2009interactive},
or marking relevant regions in images \citep{von2006peekaboom}.
But there are certain types of information which cannot be easily reduced to annotating a portion of the input, such as the absence of a certain word, or the presence of at least two words.
In this work, we tap into the power of natural language and allow annotators to provide supervision to a classifier via \emph{natural language explanations}.

Specifically, we propose a framework in which annotators provide a natural language explanation for each label they assign to an example (see Figure~\ref{fig:interface}).
These explanations are parsed into logical forms representing labeling functions (LFs), functions that heuristically map examples to labels \cite{ratner2016data}.
The labeling functions are then executed on many unlabeled examples, resulting in a large, weakly-supervised training set that is then used to train a classifier.

Semantic parsing of natural language into logical forms is recognized as a challenging problem and has been studied extensively \cite{zelle96geoquery, zettlemoyer05ccg, liang11dcs, liang2016executable}.
One of our major findings is that in our setting, even a simple rule-based semantic parser suffices for three reasons:
First, we find that the majority of incorrect LFs can be automatically filtered out either semantically
(e.g., is it consistent with the associated example?) or pragmatically (e.g., does it avoid assigning the same label to the entire training set?).
Second, LFs near the gold LF in the space of logical forms are often just as accurate (and sometimes even more accurate).
Third, techniques for combining weak supervision sources are built to tolerate some noise \cite{alfonseca2012pattern, takamatsu2012reducing, ratner2018snorkel}.
The significance of this is that we can deploy the same semantic parser across tasks without task-specific training.
We show how we can tackle a real-world biomedical application with the same semantic parser used to extract instances of spouses.

Our work is most similar to that of \citet{srivastava2017joint}, who also use natural language explanations to train a classifier, but with two important differences.
First, they jointly train a task-specific semantic parser and classifier,
whereas we use a simple rule-based parser.
In Section~\ref{sec:experiments}, we find that in our weak supervision framework, the
rule-based semantic parser and the perfect parser yield nearly identical
downstream performance.
Second, while they use the logical forms of explanations to produce \emph{features} that are fed directly to a classifier, we use them as \emph{functions} for labeling a much larger training set.
In Section~\ref{sec:experiments}, we show that using functions yields a 9.5 F1 improvement (26\% relative improvement) over features, and that the F1 score scales with the amount of available unlabeled data.

% Experiments
We validate our approach on two existing datasets from the literature (extracting spouses from news articles and disease-causing chemicals from biomedical abstracts)
and one real-world use case with our biomedical collaborators at OccamzRazor to extract protein-kinase interactions related to Parkinson's disease from text.
We find empirically that users are able to train classifiers with comparable F1 scores up to 100$\times$ faster when they provide natural language explanations instead of individual labels. Our code is available at \url{https://github.com/HazyResearch/babble}.

\begin{figure*}[t]
  \centering
  \includegraphics[width=5in]{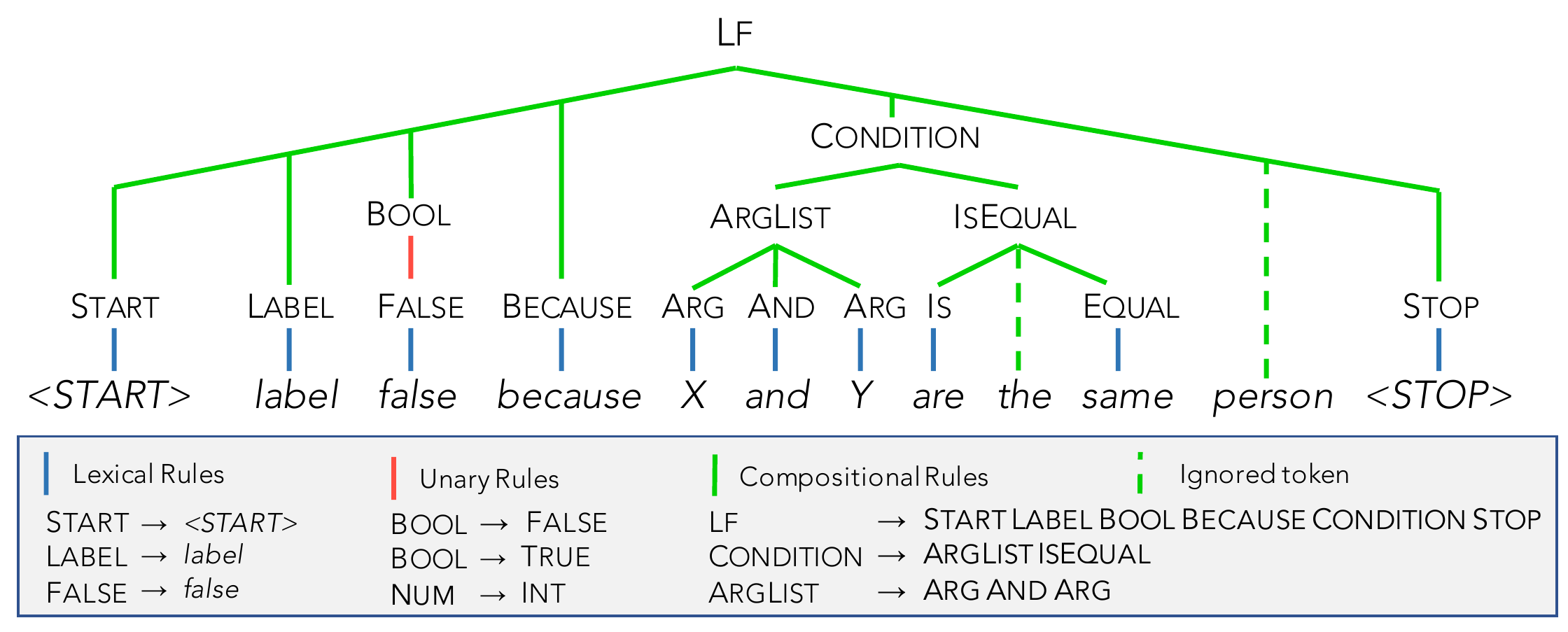}
  \caption{Valid parses are found by iterating over increasingly large subspans of the input looking for matches among the right hand sides of the rules in the grammar. Rules are either lexical (converting tokens into symbols), unary (converting one symbol into another symbol), or compositional (combining many symbols into a single higher-order symbol). A rule may optionally ignore unrecognized tokens in a span (denoted here with a dashed line).
  }
  \label{fig:semparser}
\end{figure*}

\section{The \bl Framework}
The \bl framework converts natural language explanations and unlabeled data into a noisily-labeled training set (see Figure~\ref{fig:walkthrough}).
There are three key components: a semantic parser, a filter bank, and a label aggregator.
The semantic parser converts natural language explanations into a set of logical forms representing labeling functions (LFs).
The filter bank removes as many incorrect LFs as possible without requiring ground truth labels.
The remaining LFs are applied to unlabeled examples to produce a matrix of labels.
This label matrix is passed into the label aggregator, which combines these potentially conflicting and overlapping labels into one label for each example.
The resulting labeled examples are then used to train an arbitrary discriminative model.

\subsection{Explanations}
\label{sec:explanations}
To create the input explanations, the user views a subset $S$ of an unlabeled dataset $D$ (where $|S| \ll |D|$) and provides for each input $x_i \in S$ a label $y_i$ and a natural language explanation $e_i$, a sentence explaining why the example should receive that label.
The explanation $e_i$ generally refers to specific aspects of the example
(e.g., in Figure~\ref{fig:walkthrough}, the location of a specific string ``his wife'').

\subsection{Semantic Parser}
\label{sec:semparser}

The semantic parser takes a natural language explanation $e_i$ and returns a set of LFs (logical forms or labeling functions) $\{f_1,\ldots,f_k\}$
of the form $f_i: \sX \rightarrow \{-1,0,1\}$ in a binary classification setting, with 0 representing abstention.
We emphasize that the goal of this semantic parser is \emph{not} to generate the single correct parse, but rather to have coverage over many potentially useful LFs.\footnote{Indeed, we find empirically that an incorrect LF nearby the correct one in the space of logical forms actually has higher end-task accuracy 57\% of the time (see Section~\ref{sec:utility}).}

We choose a simple rule-based semantic parser that can be used without any training.
Formally, the parser uses a set of rules of the form $\alpha \rightarrow \beta$, where $\alpha$ can be replaced by the token(s) in $\beta$ (see Figure~\ref{fig:semparser} for example rules).
To identify candidate LFs, we recursively construct a set of valid parses for each span of the explanation, based on the substitutions defined by the grammar rules.
At the end, the parser returns all valid parses (LFs in our case) corresponding to the entire explanation.

We also allow an arbitrary number of tokens in a given span to be ignored when looking for a matching rule.
This improves the ability of the parser to handle unexpected input, such as unknown words or typos,
since the portions of the input that \emph{are} parseable can still result in a valid parse.
For example, in Figure~\ref{fig:semparser}, the word ``person'' is ignored.

All predicates included in our grammar (summarized in Table~\ref{tab:grammar}) are provided to annotators, with minimal examples of each in use \mbox{(Appendix~\ref{appendix:rules})}.
Importantly, all rules are domain independent (e.g., all three relation extraction tasks that we tested used the same grammar), making the semantic parser easily transferrable to new domains.
Additionally, while this paper focuses on the task of relation extraction, in principle the \bl framework can be applied to other tasks or settings by extending the grammar with the necessary primitives (e.g., adding primitives for rows and columns to enable explanations about the alignments of words in tables).
To guide the construction of the grammar, we collected 500 explanations for the \spouse domain from workers on Amazon Mechanical Turk and added support for the most commonly used predicates.
These were added before the experiments described in Section~\ref{sec:experiments}.
The grammar contains a total of 200 rule templates.

{\renewcommand{\arraystretch}{1.3}
\begin{table}
  \footnotesize
  \centering
  \begin{tabular}{p{3cm}|p{4cm}}
    \toprule
    Predicate  & Description \\
    \midrule
    \lmtt{bool, string, int, float, tuple, list, set} &
      Standard primitive data types \\
    \lmtt{and, or, not, any, all, none} &
      Standard logic operators \\
    $=$, $\neq$, $<$, $\leq$, $>$, $\geq$ &
      Standard comparison operators \\
    \lmtt{lower, upper, capital, all\_caps} &
      Return True for strings of the corresponding case\\
    \lmtt{starts\_with, ends\_with, substring} &
      Return True if the first string starts/ends with or contains the second \\
    \lmtt{person, location, date, number, organization} &
      Return True if a string has the corresponding NER tag \\
    \lmtt{alias} &
      A frequently used list of words may be predefined and referred to with an alias \\
    \lmtt{count, contains, intersection} &
      Operators for checking size, membership, or common elements of a {\lmtt{list/set}} \\
    \lmtt{map, filter} &
      Apply a functional primitive to each member of {\lmtt{list/set}} to transform or filter the elements \\
    \lmtt{word\_distance, character\_distance} &
      Return the distance between two strings by words or characters \\
    \lmtt{left, right, between, within} &
      Return as a string the text that is left/right/within some distance of a string or between two designated strings\\
    \bottomrule
  \end{tabular}
  \caption{
    Predicates in the grammar supported by \bl's rule-based semantic parser.
  }
  \label{tab:grammar}
\end{table}
}

\subsection{Filter Bank}
\label{sec:filter}

The input to the filter bank is a set of candidate LFs produced by the semantic parser.
The purpose of the filter bank is to discard as many incorrect LFs as possible without requiring additional labels.
It consists of two classes of filters: semantic and pragmatic.

Recall that each explanation $e_i$ is collected in the context of a specific labeled example $(x_i, y_i)$. The \emph{semantic filter} checks for LFs that are inconsistent with their corresponding example;
formally, any LF $f$ for which $f(x_i) \neq y_i$ is discarded.
For example, in the first explanation in Figure~\ref{fig:walkthrough}, the word ``right'' can be interpreted as either ``immediately'' (as in ``right before'')  or simply ``to the right.''
The latter interpretation results in a function that is inconsistent with the associated example (since ``his wife'' is actually to the left of person 2), so it can be safely removed.

The \emph{pragmatic filters} removes LFs that are constant, redundant, or correlated.
For example, in Figure~\ref{fig:walkthrough}, {\lmtt{LF\_2a}} is constant, as it labels every example positively (since all examples contain two people from the same sentence).
{\lmtt{LF\_3b}} is redundant, since even though it has a different syntax tree from {\lmtt{LF\_3a}}, it labels the training set identically and therefore provides no new signal.

Finally, out of all LFs from the same explanation that pass all the other filters, we keep only the most specific (lowest coverage) LF.
This prevents multiple correlated LFs from a single example from dominating.

As we show in Section~\ref{sec:experiments}, over three tasks, the filter bank removes over 95\% of incorrect parses,
and the incorrect ones that remain have average end-task accuracy within 2.5 points of the corresponding correct parses.

\subsection{Label Aggregator}
\label{sec:aggregator}

\begin{figure*}[]
    \centering
    \includegraphics[width=6.25in]{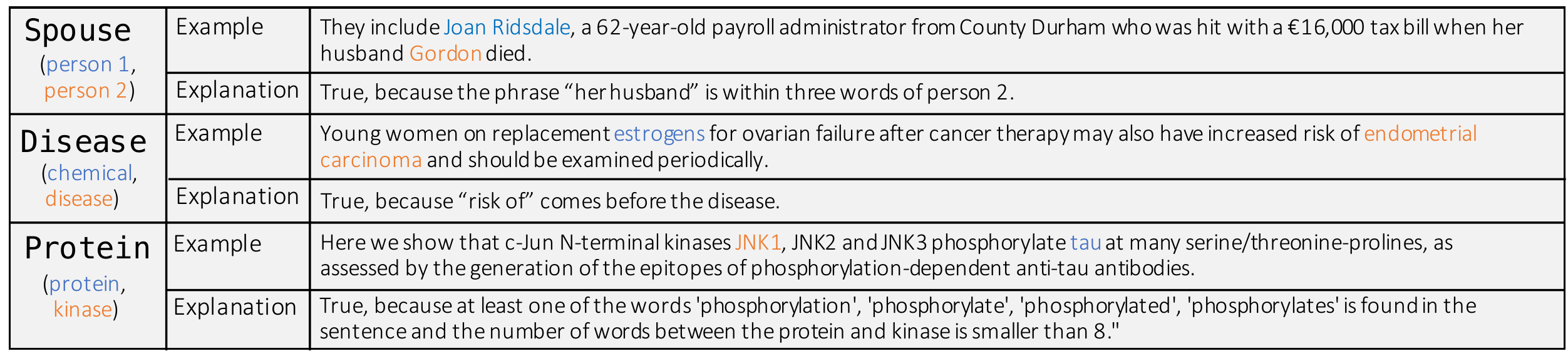}
    \caption{An example and explanation for each of the three datasets.}
    \label{fig:datasets}
\end{figure*}

The label aggregator combines multiple (potentially conflicting) suggested labels from the LFs and combines them into a single probabilistic label per example.
Concretely, if $m$ LFs pass the filter bank and are applied to $n$ examples, the label aggregator implements a function $f: \{-1,0,1\}^{m \times n} \rightarrow [0,1]^n$.

A naive solution would be to use a simple majority vote, but this fails to account for the fact that LFs can vary widely in accuracy and coverage.
Instead, we use data programming \cite{ratner2016data}, which models the relationship between the true labels and the output of the labeling functions as a factor graph.
More specifically, given the true labels $Y \in \{-1, 1\}^n$ (latent) and label matrix $\Lambda \in \{-1, 0, 1\}^{m \times n}$ (observed) where $\Lambda_{i,j} = \textrm{LF}_i(x_j)$,
we define two types of factors representing labeling propensity and accuracy:
\begin{align}
    & \phi_{i,j}^\text{Lab}(\Lambda, Y) = \mathds{1}\{\Lambda_{i,j} \neq 0\} \\
    & \phi_{i,j}^\text{Acc}(\Lambda, Y) = \mathds{1}\{\Lambda_{i,j} = y_j\}.
\end{align}
Denoting the vector of factors pertaining to a given data point $x_j$ as $\phi_j(\Lambda, Y) \in \mathbb R^m$, define the model:
\begin{align}
    p_w(\Lambda, Y) = Z_w^{-1}\exp \Big( \sum_{j=1}^n w \cdot \phi_j (\Lambda, Y) \Big),
\end{align}
where $w \in \mathbb R^{2m}$ is the weight vector and $Z_w$ is the normalization constant. To learn this model without knowing the true labels $Y$, we minimize the negative log marginal likelihood given the observed labels $\Lambda$:
\begin{align}
    \hat{w} = \argmin_w - \log \sum_Y p_w(\Lambda, Y)
\end{align}
using SGD and Gibbs sampling for inference, and then use the marginals $p_{\hat{w}}(Y \mid \Lambda)$ as probabilistic training labels.

Intuitively, we infer accuracies of the LFs based on the way they overlap and conflict with one another. Since noisier LFs are more likely to have high conflict rates with others, their corresponding accuracy weights in $w$ will be smaller, reducing their influence on the aggregated labels.

\subsection{Discriminative Model}
\label{sec:disc}
The noisy training set that the label aggregator outputs is used to train an arbitrary discriminative model.
One advantage of training a discriminative model on the task instead of using the label aggregator as a classifier directly is that the label aggregator only takes into account those signals included in the LFs.
A discriminative model, on the other hand, can incorporate features that were not identified by the user but are nevertheless informative.\footnote{We give an example of two such features in Section~\ref{sec:funcfeat}.}
Consequently, even examples for which all LFs abstained can still be classified correctly.
Additionally, passing supervision information from the user to the model in the form of a dataset---rather than hard rules---promotes generalization in the new model (rather than memorization), similar to distant supervision \cite{mintz2009distant}.
On the three tasks we evaluate, using the discriminative model averages 4.3 F1 points higher than using the label aggregator directly.

For the results reported in this paper, our discriminative model is a simple logistic regression classifier with generic features defined over dependency paths.\footnote{https://github.com/HazyResearch/treedlib}
These features include unigrams, bigrams, and trigrams of lemmas, dependency labels, and part of speech tags found in the siblings, parents, and nodes between the entities in the dependency parse of the sentence.
We found this to perform better on average than a biLSTM, particularly for the traditional supervision baselines with small training set sizes; it also provided easily interpretable features for analysis.

\begin{table}
  \centering
  \setlength\tabcolsep{7.5pt}
  \begin{tabular}{lrrrrr}
    \toprule
    Task                   & Train & Dev & Test & \% Pos.\\
    \midrule
    {\lmtt Spouse\xspace}  & 22195 & 2796  & 2697 & 8\% \\
    {\lmtt Disease\xspace} & 6667  & 773   & 4101 & 23\% \\
    {\lmtt Protein\xspace} & 5546  & 1011  & 1058 & 22\% \\
    \bottomrule
  \end{tabular}
  \caption{The total number of unlabeled training examples (a pair of annotated entities in a sentence), labeled development examples (for hyperparameter tuning), labeled test examples (for \mbox{assessment}), and the fraction of positive labels in the test split.}
  \label{tab:datasets}
\end{table}

\section{Experimental Setup}
\label{sec:setup}
We evaluate the accuracy of \bl on three relation extraction tasks, which we refer to as \spouse, \disease, and \protein.
The goal of each task is to train a classifier for predicting whether the two entities in an example are participating in the relationship of interest, as described below.

\begin{table*}[tb]
    \centering
    \setlength\tabcolsep{12pt}
    \begin{tabular}{llcllllllll}
      \toprule
                & BL   && \multicolumn{7}{c}{TS}\\
      \cmidrule{2-2} \cmidrule{4-10}
      \# Inputs & 30   && 30   & 60  &  150  & 300  & 1,000 & 3,000 & 10,000 \\
      \midrule
      \spouse   & 50.1 && 15.5 & 15.9 & 16.4 & 17.2 & 22.8 & 41.8 & 55.0  \\
      \disease  & 42.3 && 32.1 & 32.6 & 34.4 & 37.5 & 41.9 & 44.5 &  -    \\
      \protein  & 47.3 && 39.3 & 42.1 & 46.8 & 51.0 & 57.6 &  -   &  -    \\
      \midrule
      Average   & 46.6 && 28.9 & 30.2 & 32.5 & 35.2 & 40.8 & 43.2 & 55.0 \\
      \bottomrule
    \end{tabular}
    \caption{
      F1 scores obtained by a classifier trained with \bl (BL) using 30 explanations or with traditional supervision (TS) using the specified number of individually labeled examples.
      \bl achieves the same F1 score as traditional supervision while using fewer user inputs by a factor of over 5 (\protein) to over 100 (\spouse).
    }
    \label{tab:results}
\end{table*}

\subsection{Datasets}
\label{sec:datasets}
Statistics for each dataset are reported in Table~\ref{tab:datasets}, with one example and one explanation for each given in Figure~\ref{fig:datasets} and additional explanations shown in Appendix~\ref{appendix:explanations}.

In the \spouse task, annotators were shown a sentence with two highlighted names and asked to label whether the sentence suggests that the two people are spouses. Sentences were pulled from the Signal Media dataset of news articles \cite{corney2016million}. Ground truth data was collected from Amazon Mechanical Turk workers, accepting the majority label over three annotations. The 30 explanations we report on were sampled randomly from a pool of 200 that were generated by 10 graduate students unfamiliar with \bl.

In the \disease task, annotators were shown a sentence with highlighted names of a chemical and a disease and asked to label whether the sentence suggests that the chemical causes the disease. Sentences and ground truth labels came from a portion of the 2015 BioCreative chemical-disease relation dataset \cite{wei2015overview}, which contains abstracts from PubMed. Because this task requires specialized domain expertise, we obtained explanations by having someone unfamiliar with \bl translate from Python to natural language labeling functions from an existing publication that explored applying weak supervision to this task \cite{ratner2018snorkel}.

The \protein task was completed in conjunction with OccamzRazor, a neuroscience company targeting biological pathways of Parkinson's disease. For this task, annotators were shown a sentence from the relevant biomedical literature with highlighted names of a protein and a kinase and asked to label whether or not the kinase influences the protein in terms of a physical interaction or phosphorylation. The annotators had domain expertise but minimal programming experience, making \bl a natural fit for their use case.

\subsection{Experimental Settings}
\label{sec:settings}
Text documents are tokenized with spaCy.\footnote{https://github.com/explosion/spaCy}
The semantic parser is built on top of the Python-based implementation SippyCup.\footnote{https://github.com/wcmac/sippycup}
On a single core, parsing 360 explanations takes approximately two seconds.
We use existing implementations of the label aggregator, feature library, and discriminative classifier described in Sections~\ref{sec:aggregator}--\ref{sec:disc} provided by the open-source project Snorkel \citep{ratner2018snorkel}.

Hyperparameters for all methods we report were selected via random search over thirty configurations on the same held-out development set.
We searched over learning rate, batch size, L$_2$ regularization, and the subsampling rate (for improving balance between classes).\footnote{Hyperparameter ranges: learning rate (1e-2 to 1e-4), batch size (32 to 128), L$_2$ regularization (0 to 100), subsampling rate (0 to 0.5)}
All reported F1 scores are the average value of 40 runs with random seeds and otherwise identical settings.

\section{Experimental Results}
\label{sec:experiments}
We evaluate the performance of \bl with respect to its rate of improvement by number of user inputs, its dependence on correctly parsed logical forms, and the mechanism by which it utilizes logical forms.

\subsection{High Bandwidth Supervision}
\label{sec:bandwidth}

In Table~\ref{tab:results} we report the average F1 score of a classifier trained with \bl using 30 explanations or traditional supervision with the indicated number of labels.
On average, it took the same amount of time to collect 30 explanations as 60 labels.\footnote{\citet{zaidan08annotator} also found that collecting annotator rationales in the form of highlighted substrings from the sentence only doubled annotation time.}
We observe that in all three tasks, \bl achieves a given F1 score with far fewer user inputs than traditional supervision, by as much as 100 times in the case of the \spouse task.
Because explanations are applied to many unlabeled examples, each individual input from the user can implicitly contribute many (noisy) labels to the learning algorithm.

\begin{figure*}[t]
  \centering
  \includegraphics[width=6.25in]{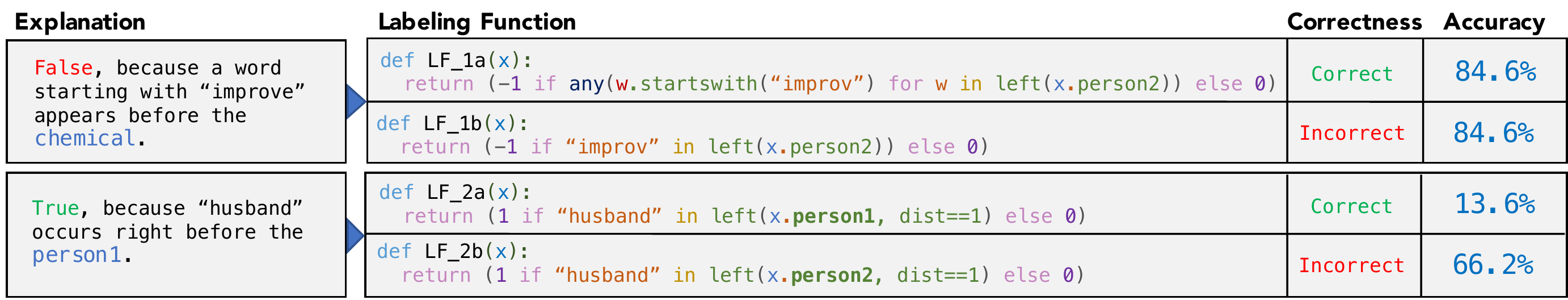}
  \caption{
    Incorrect LFs often still provide useful signal.
    On top is an incorrect LF produced for the \disease task that had the same accuracy as the correct LF.
    On bottom is a correct LF from the \spouse task and a more accurate incorrect LF discovered by randomly perturbing one predicate at a time as described in Section~\ref{sec:utility}.
    (Person 2 is always the second person in the sentence).
  }
  \label{fig:perturbations}
 \end{figure*}

We also observe, however, that once the number of labeled examples is sufficiently large, traditional supervision once again dominates, since ground truth labels are preferable to noisy ones generated by labeling functions.
However, in domains where there is much more unlabeled data available than labeled data (which in our experience is most domains), we can gain in supervision efficiency from using \bl.

Of those explanations that did not produce a correct LF, 4\% were caused by the explanation referring to unsupported concepts (e.g., one explanation referred to ``the subject of the sentence,'' which our simple parser doesn't support).
Another 2\% were caused by human errors (the correct LF for the explanation was inconsistent with the example).
The remainder were due to unrecognized paraphrases (e.g., the explanation said ``the order of appearance is X, Y'' instead of a supported phrasing like ``X comes before Y'').

\begin{table}[tb]
  \small
  \centering
  \setlength\tabcolsep{4pt}
  \begin{tabular}{@{}rrrcrrcrr@{}}
    \toprule
     & \multicolumn{2}{c}{Pre-filters} && \multicolumn{2}{c}{Discarded} && \multicolumn{2}{c}{Post-filters} \\
     \cmidrule{2-3} \cmidrule{5-6} \cmidrule{8-9}
     & LFs & Correct && Sem. & Prag. && LFs & Correct\\
    \midrule
    \spouse   & 156 & 10\% && 19 & 118 && 19 & 84\% \\
    \disease  & 102 & 23\% && 34 & 40  && 28 & 89\% \\
    \protein  & 122 & 14\% && 44 & 58  && 20 & 85\% \\
    \bottomrule
  \end{tabular}
  \caption{
    The number of LFs generated from 30 explanations (pre-filters), discarded by the filter bank, and remaining (post-filters), along with the percentage of LFs that were correctly parsed from their corresponding explanations.
  }
  \label{tab:filters}
\end{table}

\subsection{Utility of Incorrect Parses}
\label{sec:utility}
In Table~\ref{tab:filters}, we report LF summary statistics before and after filtering.
LF correctness is based on exact match with a manually generated parse for each explanation.
Surprisingly, the simple heuristic-based filter bank successfully removes over 95\% of incorrect LFs in all three tasks, resulting in final LF sets that are 86\% correct on average.
Furthermore, among those LFs that pass through the filter bank, we found that the average difference in end-task accuracy between correct and incorrect parses is less than 2.5\%.
Intuitively, the filters are effective because it is quite difficult for an LF to be parsed from the explanation, label its own example correctly (passing the semantic filter), and not label all examples in the training set with the same label or identically to another LF (passing the pragmatic filter).

\begin{table}[t]
  \small
  \centering
  \setlength\tabcolsep{15pt}
%   \begin{tabular}{@{}rrrcrrcrr@{}}
  \begin{tabular}{lccc}
    \toprule
              & BL-FB & BL & BL+PP \\
    \midrule
    \spouse   & 15.7 & 50.1 & 49.8 \\
    \disease  & 39.8 & 42.3 & 43.2 \\
    \protein  & 38.2 & 47.3 & 47.4 \\
    \midrule
    Average   & 31.2 & 46.6 & 46.8 \\
    \bottomrule
  \end{tabular}
  \caption{F1 scores obtained using \bl with no filter bank (BL-FB), as normal (BL), and with a perfect parser (BL+PP) simulated by hand.}
  \label{tab:variants}
\end{table}

We went one step further: using the LFs that would be produced by a perfect semantic parser as starting points, we searched for ``nearby'' LFs (LFs differing by only one predicate) with higher end-task accuracy on the test set and succeeded 57\% of the time (see Figure~\ref{fig:perturbations} for an example).
In other words, when users provide explanations, the signals they describe provide good starting points, but they are actually unlikely to be optimal.
This observation is further supported by Table~\ref{tab:variants}, which shows that the filter bank is necessary to remove clearly irrelevant LFs, but with that in place, the simple rule-based semantic parser and a perfect parser have nearly identical average F1 scores.

% \vspace{-20mm}
\begin{figure}[t]
  \centering
  \vspace{-6mm}
  \includegraphics[width=3in,left]{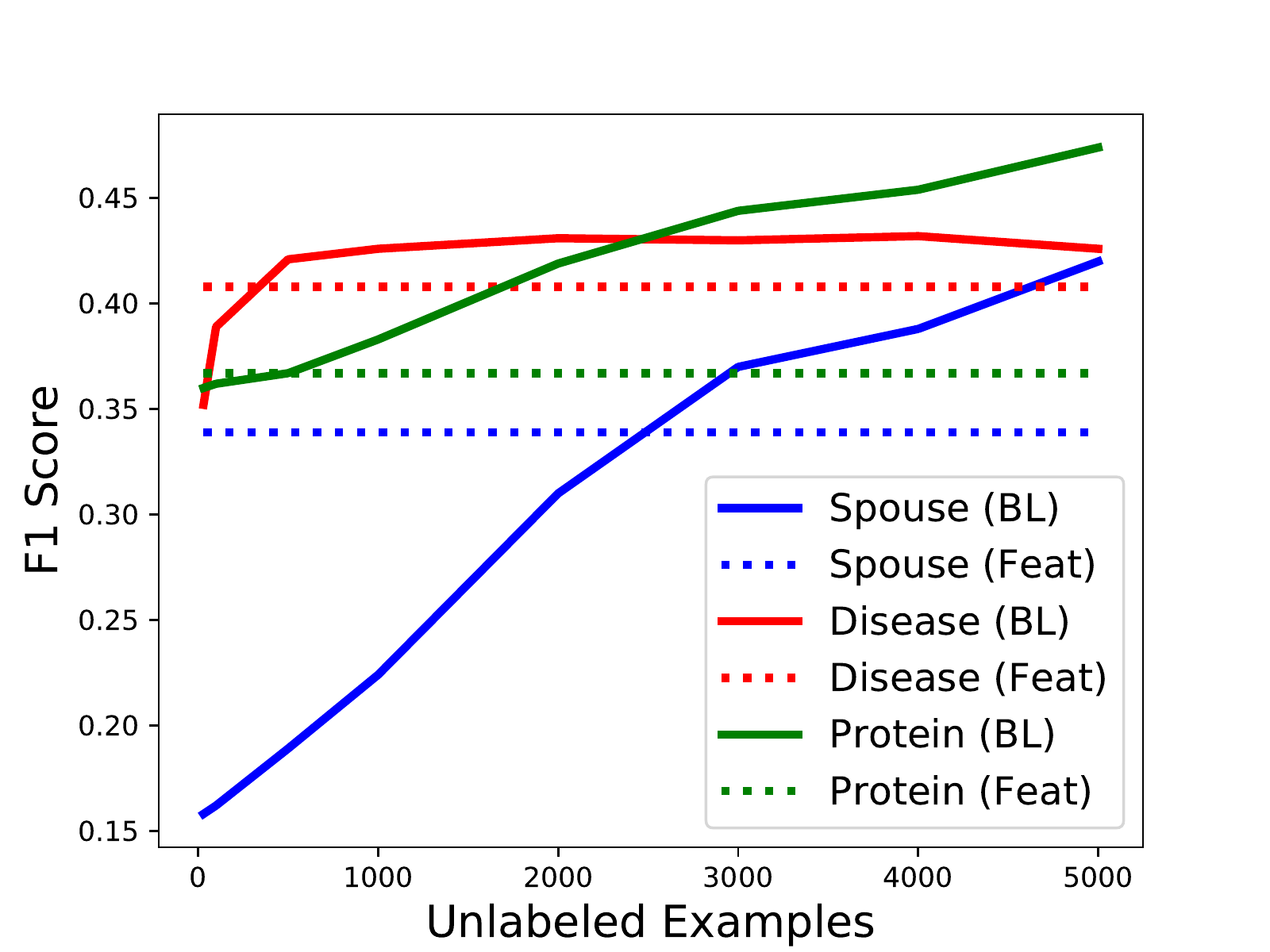}
  \caption{
    When logical forms of natural language explanations are used as functions for data programming (as they are in \bl), performance can improve with the addition of \emph{unlabeled} data,
    whereas using them as features does not benefit from unlabeled data.
  }
  \label{fig:scaling}
\end{figure}

\subsection{Using LFs as Functions or Features}
\label{sec:funcfeat}
Once we have relevant logical forms from user-provided explanations, we have multiple options for how to use them.
\citet{srivastava2017joint} propose using these logical forms as features in a linear classifier, essentially using a traditional supervision approach with user-specified features.
We choose instead to use them as functions for weakly supervising the creation of a larger training set via data programming \cite{ratner2016data}.
In Table~\ref{tab:funcorfeat}, we compare the two approaches directly, finding that the the data programming approach outperforms a feature-based one by 9.5 F1 points on average with the rule-based parser, and by 4.5 points with a perfect parser.

We attribute this difference primarily to the ability of data programming to utilize a larger feature set and unlabeled data.
In Figure~\ref{fig:scaling}, we show how the data programming approach improves with the number of unlabeled examples, even as the number of LFs remains constant.
We also observe qualitatively that data programming exposes the classifier to additional patterns that are correlated with our explanations but not mentioned directly.
For example, in the \disease task, two of the features weighted most highly by the discriminative model were the presence of the trigrams ``could produce a'' or ``support diagnosis of'' between the chemical and disease, despite none of these words occurring in the explanations for that task.
In Table~\ref{tab:funcorfeat} we see a 4.3 F1 point improvement (10\%) when we use the discriminative model that can take advantage of these features rather than applying the LFs directly to the test set and making predictions based on the label aggregator's outputs.

\begin{table}[tb]
  \small
  \centering
  \setlength\tabcolsep{4.5pt}
  \begin{tabular}{@{}lcrccrcc@{}}
    \toprule
              & BL-DM && BL & BL+PP && Feat & Feat+PP \\
    \midrule
    \spouse   & 46.5 && 50.1 & 49.8 && 33.9 & 39.2 \\
    \disease  & 39.7 && 42.3 & 43.2 && 40.8 & 43.8 \\
    \protein  & 40.6 && 47.3 & 47.4 && 36.7 & 44.0 \\
    \midrule
    Average   & 42.3 && 46.6 & 46.8 && 37.1 & 42.3 \\
    \bottomrule
  \end{tabular}
  \caption{
    F1 scores obtained using explanations as functions for data programming (BL) or features (Feat), optionally with no discriminative model \mbox{(-DM)} or using a perfect parser (+PP).
  }
  \label{tab:funcorfeat}
\end{table}

\section{Related Work and Discussion}
\label{sec:related}

% Natural language for concept learning/Learning from natural language supervision
Our work has two themes: modeling natural language explanations/instructions and learning from weak supervision.
The closest body of work is on ``learning from natural language.''
As mentioned earlier, \citet{srivastava2017joint}
convert natural language explanations into classifier features (whereas we convert them into labeling functions).
\citet{goldwasser11instructions} convert natural language into concepts (e.g., the rules of a card game).
\citet{ling2017teaching} use natural language explanations to assist in supervising an image captioning model.
\citet{weston2016dialog,li2016learning} learn from natural language feedback in a dialogue.
\citet{wang2017naturalizing} convert natural language definitions to rules in a semantic parser to build up progressively higher-level concepts.

% Learning semantic parsers from weak supervision
We lean on the formalism of semantic parsing \citep{zelle96geoquery,zettlemoyer05ccg, liang2016executable}.
One notable trend is to learn semantic parsers from weak supervision \citep{clarke10world,liang11dcs},
whereas our goal is to obtain weak supervision signal from semantic parsers.

% Learning from weak supervision
The broader topic of weak supervision has received much attention;
we mention some works most related to relation extraction.
In distant supervision \cite{craven1999constructing,mintz2009distant} and
multi-instance learning \cite{riedel2010modeling,hoffmann2011knowledge},
an existing knowledge base is used to (probabilistically) impute a training set.
Various extensions have focused on aggregating a variety of supervision sources
by learning generative models from noisy labels \cite{alfonseca2012pattern,
takamatsu2012reducing, roth2013combining, ratner2016data, varma2017socratic}.

% Natural language explanations
Finally, while we have used natural language explanations as \emph{input} to train models,
they can also be \emph{output} to interpret models \cite{krening2017learning, lei2016rationalizing}.
More generally, from a machine learning perspective, labels are the primary asset, but they are a low bandwidth signal between annotators and the learning algorithm.
Natural language opens up a much higher-bandwidth communication channel.
We have shown promising results in relation extraction (where one explanation can be ``worth'' 100 labels),
and it would be interesting to extend our framework to other tasks and more interactive settings.

\section*{Reproducibility}
The code, data, and experiments for this paper are available on the CodaLab platform at
{\small \url{https://worksheets.codalab.org/worksheets/0x900e7e41deaa4ec5b2fe41dc50594548/}}.\\

\noindent
Refactored code with simplified dependencies, performance and speed improvements, and interactive tutorials can be found on Github: \\
{\small \url{https://github.com/HazyResearch/babble}}.

\section*{Acknowledgments}
We gratefully acknowledge the support of the following organizations:
DARPA under No. N66001-15-C-4043 (SIMPLEX), No. FA8750-17-2-0095 (D3M), No. FA8750-12-2-0335 (XDATA), and No. FA8750-13-2-0039 (DEFT),
DOE under No. 108845,
NIH under No. U54EB020405 (Mobilize),
ONR under No. N000141712266 and No. N000141310129,
AFOSR under No. 580K753,
the Intel/NSF CPS Security grant No. 1505728,
the Michael J. Fox Foundation for Parkinson’s Research under Grant No. 14672, % Martin
the Secure Internet of Things Project,
Qualcomm,
Ericsson,
Analog Devices,
the Moore Foundation,
the Okawa Research Grant,
American Family Insurance,
Accenture,
Toshiba,
the National Science Foundation Graduate Research Fellowship under Grant No. DGE-114747, % Braden, Paroma
the Stanford Finch Family Fellowship, % Braden
the Joseph W. and Hon Mai Goodman Stanford Graduate Fellowship, % Paroma
an NSF CAREER Award IIS-1552635, % Percy
and the members of the Stanford DAWN project: Facebook, Google, Intel, Microsoft, NEC, Teradata, and VMware.

We thank Alex Ratner for his assistance with data programming, Jason Fries and the many members of the Hazy Research group and Stanford NLP group who provided feedback and tested early prototyptes, Kaya Tilev from the Stanford Graduate School of Business for helpful discussions early on, and the OccamzRazor team: Tarik Koc, Benjamin Angulo, Katharina S. Volz, and Charlotte Brzozowski.

The U.S. Government is authorized to reproduce and distribute reprints for
Governmental purposes notwithstanding any copyright notation thereon. Any opinions,
findings, and conclusions or recommendations expressed in this material are those of
the authors and do not necessarily reflect the views, policies, or endorsements, either
expressed or implied, of DARPA, DOE, NIH, ONR, AFOSR, NSF, or the U.S. Government.

\bibliography{babble_references}
\bibliographystyle{acl_natbib}

\newpage
\onecolumn
\appendix

\section{Predicate Examples}
\label{appendix:rules}

{\normalsize{

Below are the predicates in the rule-based semantic parser grammar, each of which may have many supported paraphrases, only one of which is listed here in a minimal example.\\

{\lmtt

\noindent \textbf{Logic} \\
and: X is true and Y is true \\
or: X is true or Y is true \\
not: X is not true \\
any: Any of X or Y or Z is true \\
all: All of X and Y and Z are true \\
none: None of X or Y or Z is true \\

\vspace{-0.16cm}
\noindent \textbf{Comparison} \\
$=$: X is equal to Y \\
$\neq$: X is not Y \\
$<$: X is smaller than Y \\
$\leq$: X is no more than Y \\
$>$: X is larger than Y \\
$\geq$: X is at least Y \\

\vspace{-0.16cm}
\noindent \textbf{Syntax} \\
lower: X is lowercase \\
upper: X is upper case \\
capital: X is capitalized \\
all\_caps: X is in all caps \\
starts\_with: X starts with "cardio" \\
ends\_with: X ends with "itis" \\
substring: X contains "-induced" \\

\vspace{-0.16cm}
\noindent \textbf{Named-entity Tags} \\
person: A person is between X and Y \\
location: A place is within two words of X \\
date: A date is between X and Y \\
number: There are three numbers in the sentence\\
organization: An organization is right after X \\

\vspace{-0.16cm}
\noindent \textbf{Lists} \\
list: (X, Y) is in Z \\
set: X, Y, and Z are true  \\
count: There is one word between X and Y \\
contains: X is in Y \\
intersection: At least two of X are in Y \\
map: X is at the start of a word in Y \\
filter: There are three capitalized words to the left of X \\
alias: A spouse word is in the sentence {\fontfamily{cmr}\selectfont
(``spouse'' is a predefined list from the user)
}\\

\vspace{-0.16cm}
\noindent \textbf{Position} \\
word\_distance: X is two words before Y \\
char\_distance: X is twenty characters after Y \\
left: X is before Y \\
right: X is after Y \\
between: X is between Y and Z \\
within: X is within five words of Y

}}}

\newpage
\onecolumn
\section{Sample Explanations}
\label{appendix:explanations}
\normalsize{
The following are a sample of the explanations provided by users for each task. \\

\noindent \textbf{\spouse} \\
\noindent Users referred to the first person in the sentence as ``X'' and the second as ``Y''. \\

{\fontfamily{pcr}\selectfont

    \noindent Label true because "and" occurs between X and Y and "marriage" occurs one word after person1. \\

    \noindent Label true because person Y is preceded by `beau'. \\

    \noindent Label false because the words "married", "spouse", "husband", and "wife" do not occur in the sentence. \\

    \noindent Label false because there are more than 2 people in the sentence and "actor" or "actress" is left of person1 or person2. \\
}

\noindent \textbf{\disease} \\

{\fontfamily{pcr}\selectfont

    \noindent Label true because the disease is immediately after the chemical and 'induc' or 'assoc' is in the chemical name.\\

    \noindent Label true because a word containing 'develop' appears somewhere before the chemical,
    and the word 'following' is between the disease and the chemical. \\

    \noindent Label true because "induced by", "caused by", or "due to" appears between the chemical and the disease." \\

    \noindent Label false because "none", "not", or "no" is within 30 characters to the left of the disease. \\

    \noindent \textbf{\protein} \\

{\fontfamily{pcr}\selectfont

    \noindent Label true because "Ser" or "Tyr" are within 10 characters of the protein. \\

    \noindent Label true because the words "by" or "with" are between the protein and kinase and the words "no", "not" or "none" are not in between the protein and kinase and the total number of words between them is smaller than 10. \\

    \noindent Label false because the sentence contains "mRNA", "DNA", or "RNA". \\

    \noindent Label false because there are two "," between the protein and the kinase with less than 30 characters between them. \\
}
}

\end{document}